\documentclass[runningheads]{llncs}
\usepackage[T1]{fontenc}
\usepackage{graphicx}
\usepackage{amsmath}
\usepackage{mathrsfs}
\usepackage{booktabs}   
\usepackage{multirow}   
\usepackage{caption}    
\begin{document}
\title{ACQ: A Deployed Two-Stage Framework for Automated Creative Quota Allocation in Large-Scale Online Advertising}
\titlerunning{ACQ: A Deployed Two-Stage Framework for Creative Allocation}
%
\author{Ruizhi Wang \and Yu Rong \and Kai Liu \and Bingjie Li \and Qingpeng Cai\thanks{Corresponding author.} \and Fei Pan \and Peng Jiang}

\authorrunning{R. Wang et al.}

\institute{Huazhong University of Science and Technology, China \\
\email{wangruizhi@hust.edu.cn}
\and
Kuaishou Technology, China \\
\email{rongyu03@kuaishou.com},
\email{liukai10@kuaishou.com},
\email{libingjie03@kuaishou.com},
\email{caiqingpeng@kuaishou.com},
\email{panfei05@kuaishou.com},
\email{jiangpeng@kuaishou.com}
}

\maketitle

\begin{abstract}
In digital advertising, demand-side platforms (DSPs) allow advertisers to create multiple ad creatives from a single photo for real-time bidding. While increasing the number of creatives can improve bidding opportunities, it cannot scale indefinitely, and the incremental advertising revenue typically exhibits diminishing returns as more creatives are generated. This raises a practical problem for DSPs: how to automatically determine an appropriate creative quota for each photo at scale. To address this problem, we propose Automated Creatives Quota (ACQ), a deployed two-stage framework for creative quota allocation in large-scale online advertising. In the first stage, ACQ predicts quota-conditioned expected revenue using a multi-task model built on an unbalanced binary tree, which is designed to handle the highly skewed revenue distribution across quota levels. In the second stage, ACQ formulates quota allocation under global capacity constraints as a multiple-choice knapsack problem (MCKP) and solves it with an efficient dual-based algorithm. Extensive offline experiments and online experiments on Kuaishou's advertising delivery platform demonstrate the effectiveness of ACQ, achieving a 6.20\% increase in platform advertising revenue.

\end{abstract}
\section{Introduction}
\begin{figure}[t]
  \centering
  \includegraphics{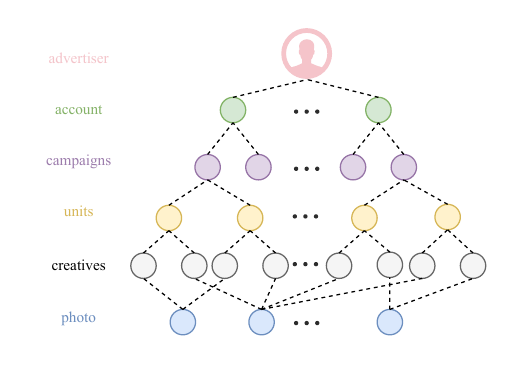}
   \caption{Hierarchical structure in an advertising system. From top to bottom, the hierarchy includes advertiser, account, campaign, unit, and creative. Each creative is generated from a single photo, while one photo can be expanded into multiple creatives by combining different titles or themes.}
  \label{fig:creative.drawio.png}
\end{figure}

Online advertising is a major revenue source for Internet companies~\cite{choi2020online}. To improve campaign effectiveness and reduce manual operations, major platforms such as Google, Amazon, Tencent, ByteDance and Kuaishou have developed sophisticated demand-side platforms (DSPs) for strategy automation, ad delivery optimization, and real-time bidding~\cite{chen2021adversarial,guo2021we}. In modern advertising systems, advertisers typically manage three hierarchical levels within an account---campaigns, units, and creatives---and must decide how many creatives should be generated for each photo. As illustrated in Figure~\ref{fig:creative.drawio.png}, a single photo can be used to generate multiple creatives by combining different titles, themes, or templates.

Generating multiple creatives from the same photo can improve bidding opportunities because different creatives may appeal to different audiences~\cite{azimi2012impact}. However, assigning more creatives to a photo does not lead to unlimited gains. From the DSP perspective, two practical issues must be considered. First, the number of creatives cannot grow indefinitely because platform resources are finite, and creating too many creatives may aggravate data sparsity, since many newly generated creatives receive little or no traffic. Second, the incremental advertising revenue brought by additional creatives usually diminishes as the creative quota increases, due to audience overlap and competition among similar creatives. Therefore, an important practical problem is to automatically determine an appropriate creative quota for each photo.

In this paper, we study the \textit{creative quota allocation problem}, which aims to optimize the number of creatives generated from each photo so as to maximize total advertising revenue under platform constraints. This problem is challenging for three reasons. First, the target variable is extremely imbalanced: on any given day, more than 90\% of photos receive zero ad revenue, which makes direct prediction difficult. Second, standard regression models do not explicitly capture the structured relationship between ad revenue and creative quota. In practice, this relationship is expected to satisfy several intuitive properties, including monotonicity, submodularity, and local smoothness. Third, the allocation problem must be solved at very large scale, involving on the order of ten million photos, which places substantial demands on memory usage and computational efficiency.

To address these challenges, we propose a two-stage framework called \textbf{Automated Creatives Quota (ACQ)}. The first stage learns the expected ad revenue of each photo under different creative quotas from historical data. To handle the severe target imbalance while preserving the structured relationship across quota levels, we develop a multi-task prediction model based on an unbalanced binary tree. The second stage allocates creative quotas under business constraints based on the predicted ad revenue. We formulate this allocation stage as a large-scale optimization problem and design an efficient solver based on Lagrangian optimization and bisection, making ACQ practical for industrial deployment.

Our main contributions are summarized as follows:
\begin{itemize}
  \item We formulate a practical and commercially important \textbf{creative quota allocation problem} for automated ad creative creation and deactivation in DSP systems.
  \item We propose ACQ, a two-stage ``prediction + allocation'' framework that jointly models quota-dependent expected ad revenue and large-scale quota allocation.
  \item We develop a multi-task prediction model based on an unbalanced binary tree to address severe target imbalance while encoding desirable structural properties, including monotonicity, submodularity, and smoothness.
 \item We design an efficient large-scale allocation solver based on Lagrangian optimization with bisection, enabling deployment on Kuaishou for problems involving tens of millions of candidate photos.
    \item We validate ACQ through extensive offline experiments on real-world data and online A/B testing in production on Kuaishou's advertising delivery platform, where it consistently outperforms strong baselines.
\end{itemize}

The rest of this paper is organized as follows. Section~2 reviews related work. Section~3 formulates the creative quota allocation problem. Section~4 presents the proposed ACQ framework. Section~5 reports the datasets and experimental results. Section~6 concludes the paper.

\section{Related Work}

\paragraph{Creative Automation in Digital Advertising.}
Digital advertising is a major application of machine learning, and creative production is a key factor in campaign performance~\cite{chen2019understanding,ford2023ai}. Existing AI applications in the creative pipeline mainly include dynamic creative optimization (DCO) and programmatic advertising creation (PAC). DCO focuses on adapting or selecting creative content according to user behavior and contextual signals~\cite{baardman2021dynamic,koren2020dynamic}, while PAC emphasizes the automated generation of ad creatives from large material libraries~\cite{guo2021vinci}. However, these studies mainly address creative generation or selection quality, rather than deciding how many creatives should be generated from a single photo under platform constraints. Our work instead studies creative quota allocation.

\paragraph{Imbalanced and Structured Prediction.}
Learning from highly imbalanced targets remains challenging for deep models~\cite{yang2021delving,deAlvis2024survey}. Existing approaches include resampling, reweighting, multi-stage training, and target transformation~\cite{zhang2021disalign,zhan2022deconfounding,sun2024cread}. Tree-structured decomposition has also shown promise for handling ordered or imbalanced prediction targets~\cite{lin2023tree}. In many practical decision systems, predictive functions are further expected to satisfy structural properties such as monotonicity, smoothness, and submodularity~\cite{wang2023multi,sun2024end}. Prior work has incorporated such properties through parametric function design, order-based losses, or constrained architectures~\cite{liu2019graph,zhao2019unified,chen2021adversarial,chin2018gaussian,wu2024cost}. Our setting combines both challenges: the target is highly imbalanced, while the quota-conditioned ad-revenue function should also satisfy meaningful structural constraints.

\paragraph{Large-Scale Constrained Allocation.}
The second stage of our framework is related to large-scale constrained decision-making, especially multiple-choice knapsack optimization and its Lagrangian relaxation~\cite{kellerer2004multiple}. Prior work has applied this formulation to industrial allocation tasks such as budget allocation and constrained decision systems~\cite{li2020spending,ai2022lbcf,zhao2019unified,wang2023multi}. Efficient dual optimization methods, including gradient-based updates and bisection-based solvers, have made such approaches practical at scale~\cite{zhao2019unified,wang2023multi}. Our work builds on this line of research but focuses on a different application: allocating creative quotas for millions of photos based on predicted quota-dependent ad revenue.


\begin{figure}[t]
  \centering
  \includegraphics[width=\textwidth]{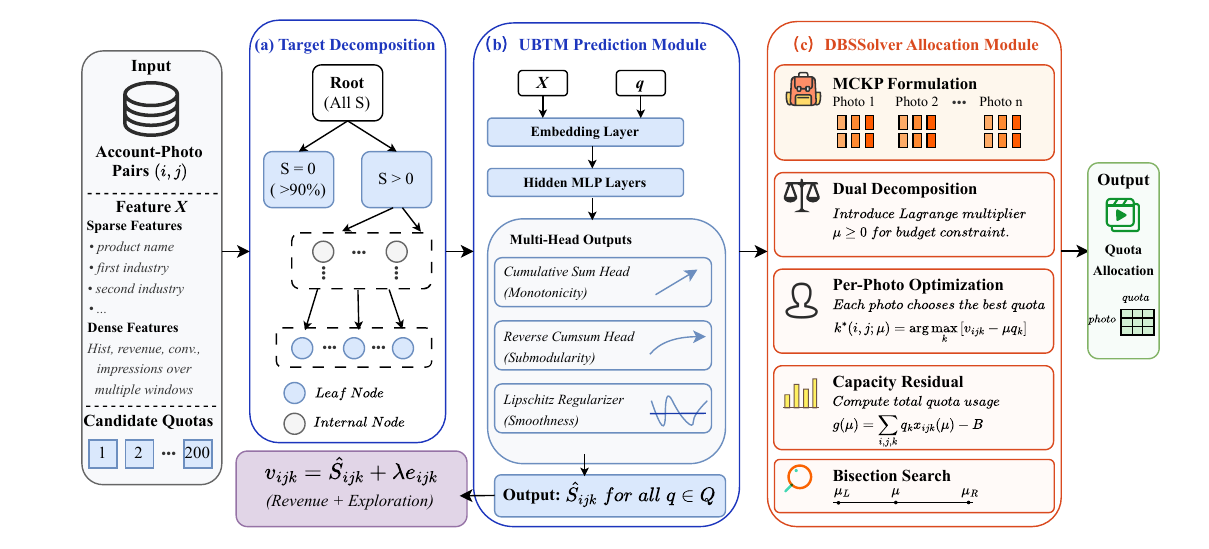}
    \caption{Overview of ACQ. The framework consists of a prediction module (UBTM) for quota-conditioned ad-revenue estimation and an allocation module (DBSSolver) for large-scale creative quota allocation.}
  \label{fig:model.drawio.png}
\end{figure}

\section{Problem Formulation}

We formulate the \textit{creative quota allocation problem} as a multiple-choice knapsack problem (MCKP). For each account-photo pair, the platform must choose one creative quota, i.e., the number of creatives to be generated from that photo. The goal is to maximize the total expected advertising revenue under a global capacity constraint on the total number of creatives.

Let $(i,j)$ denote an account-photo pair, where $i \in \{1,\dots,I\}$ indexes advertiser accounts and $j \in \{1,\dots,J_i\}$ indexes photos under account $i$. For each pair $(i,j)$, we consider a set of candidate creative quota levels $\mathcal{Q}=\{q_1,q_2,\dots,q_K\}$, where $q_k$ is the number of creatives assigned under option $k$. In our production system, the maximum quota is 200, so $\mathcal{Q}=\{1,2,\dots,200\}$.

For each account-photo pair $(i,j)$ and quota option $q_k$, let $x_{ijk}\in\{0,1\}$ be a binary decision variable, where $x_{ijk}=1$ indicates that quota $q_k$ is selected for pair $(i,j)$, and $x_{ijk}=0$ otherwise. Let $v_{ijk}$ denote the allocation value of assigning quota $q_k$ to pair $(i,j)$. The optimization problem is formulated as
\begin{equation}
\begin{aligned}
\max \quad & \sum_{i=1}^{I} \sum_{j=1}^{J_i} \sum_{k=1}^{K} v_{ijk} x_{ijk} \\
\text{s.t.} \quad & \sum_{i=1}^{I} \sum_{j=1}^{J_i} \sum_{k=1}^{K} q_k x_{ijk} \leq B, \\
& \sum_{k=1}^{K} x_{ijk} = 1, \quad \forall i,j, \\
& x_{ijk} \in \{0,1\}, \quad \forall i,j,k,
\end{aligned}
\label{eq:problem}
\end{equation}
where $B$ is the global capacity budget for the total number of creatives. The first constraint limits the total number of allocated creatives, while the second ensures that exactly one quota level is selected for each account-photo pair.

The allocation value $v_{ijk}$ is defined as
\begin{equation}
v_{ijk} = \hat{S}_{ijk} + \lambda \, e_{ijk},
\label{eq:rijk}
\end{equation}
where $\hat{S}_{ijk}$ is the predicted expected ad revenue for pair $(i,j)$ under quota $q_k$, produced by the first-stage prediction model, and $e_{ijk}$ is an exploration score. The exploration score is constructed based on the upper confidence bound (UCB) principle~\cite{auer2002finite} to balance exploitation and exploration in deployment. The coefficient $\lambda \ge 0$ controls the strength of the exploration term.

\section{Methodology}

In this section, we present the details of ACQ, as illustrated in Figure \ref{fig:model.drawio.png}. In the prediction module, we address the challenge of unbalanced target variables and examine network properties to ensure monotonicity, submodularity, and smoothness. In the allocation module, we tackle the large-scale allocation problem using a Lagrangian dual optimization method with bisection.

\subsection{Prediction Module}

\subsubsection{Modeling with an unbalanced binary tree}

The prediction module estimates the expected ad revenue of an account-photo pair under different creative quotas. Let the training set be $\mathcal{D}=\{(X_t, q_t, S_t)\}_{t=1}^{m}$, where $m$ denotes the total number of samples in the dataset, $X_t$ denotes the input features, $q_t$ the creative quota, and $S_t$ the observed ad revenue. The goal of the first stage is to learn a quota-conditioned predictor $\hat{S}(X,q)$. A major challenge is that the target distribution is highly imbalanced: many samples have zero ad revenue, while the positive values are heavy-tailed, making standard regression unstable on low-frequency but high-value regions~\cite{yang2021delving}.

To address this issue, we propose an \textbf{Unbalanced Binary Tree Model (UBTM)}. The key idea is to decompose the ad-revenue target space into a hierarchy of binary classification tasks, which serves as auxiliary supervision in a multi-task framework. This design is inspired by tree-based neural models such as TDM~\cite{zhu2018learning} and TPM~\cite{lin2023tree}, which are effective for skewed data distributions.

Specifically, we construct an unbalanced binary tree $T$ over the target variable $S$. The root node covers the full support of ad revenue values. Its left child is a singleton node corresponding to zero ad revenue, while the right subtree recursively partitions the positive target region. To better fit the long-tail distribution, the leaf nodes in the positive subtree are built by equal-frequency partitioning over samples with $S>0$. Thus, each node corresponds to either a singleton set or an interval of ad revenue values.

Suppose the tree has $N$ nodes and $N_f$ leaf nodes. Each internal node is associated with a binary classifier, resulting in $N-N_f$ auxiliary classification tasks. For any node $\mathcal{N}_k$, let $p_k=\{\mathcal{N}_k^{(0)},\mathcal{N}_k^{(1)},\dots,\mathcal{N}_k^{(d(k))}\}$ denote the path from the root to $\mathcal{N}_k$, where $d(k)$ is its depth. Then the probability that the target falls into node $\mathcal{N}_k$ can be factorized as
\begin{equation}
P(S \in \mathcal{N}_k \mid X,q,T)
=
\prod_{t=1}^{d(k)}
P\!\left(
S \in \mathcal{N}_k^{(t)}
\mid
S \in \mathcal{N}_k^{(t-1)}, X,q,T
\right).
\label{eq:tree_prob}
\end{equation}

For a leaf node, this probability represents the likelihood that the target ad revenue falls into the corresponding interval.

\subsubsection{Multi-task learning}

We train UBTM in a multi-task manner~\cite{zhang2018overview}, where the main task is quota-conditioned ad-revenue regression and the auxiliary tasks are derived from the internal nodes of the unbalanced binary tree. This design allows the model to exploit hierarchical supervision over the highly imbalanced target space and improves generalization across different regions of the ad-revenue distribution.

For the training set $\mathcal{D} = \{(X_t, q_t, S_t)\}_{t=1}^m$, the target value $S_t$ of each sample $t$ is assigned to a leaf node, which uniquely determines a root-to-leaf path. Each internal node on this path defines a binary auxiliary classification label. Suppose the tree contains $N-N_f$ internal nodes. For an auxiliary task $a \in \{1,\dots,N-N_f\}$, let $z_{t,a} \in \{0,1\}$ be the binary label for sample $t$, $\hat{p}_{t,a}$ be the predicted conditional probability, and $m_{t,a} \in \{0,1\}$ be an indicator of whether node $a$ lies on the path of sample $t$. The auxiliary classification loss is defined as:
\begin{equation}
\mathcal{L}_{\text{tree}}
=
\frac{1}{m}
\sum_{t=1}^{m}
\sum_{a=1}^{N-N_f}
m_{t,a}
\left[
-z_{t,a}\log \hat p_{t,a}
-(1-z_{t,a})\log(1-\hat p_{t,a})
\right].
\label{eq:l1}
\end{equation}

In addition to hierarchical classification, we use a regression head to predict the observed ad revenue. Let $\hat S_t$ denote the predicted ad revenue for sample $t$. The main regression loss is
\begin{equation}
\mathcal{L}_{\text{reg}}
=
\frac{1}{m}
\sum_{t=1}^{m}
(S_t-\hat S_t)^2.
\label{eq:l3}
\end{equation}

We further introduce an auxiliary uncertainty term following prior work on tree-based prediction~\cite{lin2023tree}. Specifically, we define 
\begin{equation}
\sigma_t
=
\sqrt{
{E}(S^2 \mid X_t,q_t,T)
-
{E}(S \mid X_t,q_t,T)^2
},
\label{eq:l2}
\end{equation}
which estimates the conditional standard deviation induced by the tree-based output distribution. The average uncertainty loss is then defined as:
\begin{equation}
\mathcal{L}_{\text{unc}}=\frac{1}{m}\sum_{t=1}^{m}\sigma_t.
\label{eq:l_unc}
\end{equation}
This term provides a measure of predictive uncertainty and serves as an auxiliary regularizer during training.

The final training objective is
\begin{equation}
\mathcal{L}
=
\alpha_1 \mathcal{L}_{\text{tree}}
+
\alpha_2 \mathcal{L}_{\text{unc}}
+
\alpha_3 \mathcal{L}_{\text{reg}},
\label{eq:total_loss}
\end{equation}
where $\mathcal{L}_{\text{tree}}$ and $\mathcal{L}_{\text{reg}}$ denote the tree-based loss and regularization term, respectively. In our experiments, we set $\alpha_1=1$, $\alpha_2=1$, and $\alpha_3=0.2$. Although adaptive weighting methods are available~\cite{kendall2018multi}, we found the model to be relatively stable under moderate variations of these coefficients in offline tuning.

\subsubsection{Modeling Structured Properties}

In this section, we model the functional relationship between the creative quota of a photo and its corresponding ad revenue. For each sample $(X,q,S)$, $X$ contains both sparse and dense features, $q$ denotes the creative quota, and $S$ is the observed ad revenue. Our goal is to learn a quota-conditioned response function $f(X,q) \approx {E}[S\mid X,q]$, while enforcing several structural properties that are consistent with domain knowledge.

For a given photo $P$, let $f_P(q)$ denote its expected ad revenue under creative quota $q$. We define the photo-level ad revenue under quota $q$ as the total revenue accumulated over the creatives generated from that photo. Based on the serving mechanism of advertising systems, where bidding and delivery are performed at the creative level, this aggregated quantity provides a natural target for quota allocation.

\paragraph{\textbf{Monotonicity.}}
A larger creative quota should not reduce the expected ad revenue of the same photo. Therefore, for any $q_1<q_2$, we expect $f_P(q_1) \le f_P(q_2)$.

To enforce this property, we adopt a multi-head architecture. The first output head predicts the base ad revenue under quota $1$, while each subsequent head predicts a non-negative increment relative to the previous quota level. Formally, let $g_1(X)$ be the output of the first head and let $g_k(X)$, $k\ge2$, be the raw outputs of the increment heads. We define the prediction under quota $k$ as
\begin{equation}
\hat{S}(X,k)=
\begin{cases}
g_1(X), & k=1,\\[4pt]
g_1(X)+\sum_{t=2}^{k}\phi\!\big(g_t(X)\big), & k>1,
\end{cases}
\label{eq:mono}
\end{equation}
where $\phi(\cdot)$ is a non-negative activation function. In our implementation, we use $\phi(z)=z^2$, so that every increment is non-negative by construction. As a result, $\hat{S}(X,k)$ is monotone non-decreasing with respect to the quota level $k$.

\paragraph{\textbf{Submodularity.}}

In practice, adding more creatives to the same photo usually yields smaller and smaller gains because different creatives often compete for overlapping audiences. Therefore, besides monotonicity, we also expect the quota-response function to exhibit diminishing marginal returns. Formally, for a given photo $P$, define the marginal gain at quota level $k$ as
\[
\Delta_P(k)=f_P(k)-f_P(k-1), \qquad k\ge 2.
\]
We expect
\[
\Delta_P(k+1) \le \Delta_P(k), \qquad \forall k\ge 2.
\]

To enforce this structure, we parameterize the quota-dependent increments in reverse order. Let $d_k(X)$ denote the increment from quota $k-1$ to quota $k$. We require
\[
d_2(X)\ge d_3(X)\ge \cdots \ge d_K(X)\ge 0.
\]
Specifically, let $h_k(X)$ be the raw output of the $k$-th decrement head and let $\psi(\cdot)$ be a non-negative activation function. We define
\begin{equation}
d_K(X)=\psi(h_K(X)),
\label{eq:dk_last}
\end{equation}
and for $2 \le k < K$,
\begin{equation}
d_k(X)=d_{k+1}(X)+\psi(h_k(X)).
\label{eq:dk_reverse}
\end{equation}
Since $\psi(h_k(X)) \ge 0$, the resulting increments are non-increasing as the quota level grows. The final prediction is then
\begin{equation}
\hat{S}(X,k)=\hat{S}(X,1)+\sum_{t=2}^{k} d_t(X), \qquad k\ge2,
\label{eq:submodular_pred}
\end{equation}
which guarantees both monotonicity and submodularity by construction.

\paragraph{\textbf{Smoothness.}}
We also encourage the learned response function to vary smoothly. This is desirable in practice because abrupt changes across neighboring quota levels often indicate overfitting or unstable extrapolation. Instead of imposing a hard smoothness constraint on the discrete quota-response curve, we regularize the Lipschitz constant of the network following prior work on Lipschitz-constrained neural models~\cite{szegedy2013intriguing,gouk2021regularisation}.

Let $z_j$ denote the spectral norm of the weight matrix in the $j$-th fully connected layer, and let $\mathcal{L}_{\text{base}}$ denote the original training objective. We define the Lipschitz regularization term as
\begin{equation}
\mathcal{L}_{\text{lip}}
=
\prod_{j=1}^{J}
\mathrm{softplus}(z_j),
\label{eq:lip}
\end{equation}
where $J$ is the number of fully connected layers and $\mathrm{softplus}(z)=\log(1+e^z)$ ensures non-negativity. The final smoothness-regularized objective is
\begin{equation}
\mathcal{L}_{\text{smoo}}
=
\mathcal{L}
+
\lambda_{\text{lip}} \mathcal{L}_{\text{lip}},
\label{eq:lip_total}
\end{equation}
where $\lambda_{\text{lip}}$ controls the strength of the regularization. This regularizer does not strictly enforce smoothness across quota levels, but it improves the stability of the learned mapping and empirically leads to smoother quota-response predictions.

\subsection{Allocation Module}

The creative quota allocation problem defined in Eq.~\ref{eq:problem} is a multiple-choice knapsack problem (MCKP), which is combinatorial and difficult to solve directly at very large scale. To obtain an efficient industrial solver, we first relax the binary decision variables to the continuous interval $[0,1]$. The relaxed problem is
\begin{equation}
\begin{aligned}
\max \quad & \sum_{i=1}^{I} \sum_{j=1}^{J_i} \sum_{k=1}^{K} v_{ijk} x_{ijk} \\
\text{s.t.} \quad & \sum_{i=1}^{I} \sum_{j=1}^{J_i} \sum_{k=1}^{K} q_k x_{ijk} \leq B, \\
& \sum_{k=1}^{K} x_{ijk} = 1, \quad \forall i,j, \\
& 0 \le x_{ijk} \le 1, \quad \forall i,j,k.
\end{aligned}
\label{eq:problem2}
\end{equation}

Since Eq.~\ref{eq:problem2} is a feasible linear program, strong duality holds, and the KKT conditions are necessary and sufficient for optimality. Introducing a nonnegative dual variable $\mu \ge 0$ for the global capacity constraint, we obtain the Lagrangian
\begin{equation}
\mathcal{L}(X,\mu)
=
\sum_{i=1}^{I} \sum_{j=1}^{J_i} \sum_{k=1}^{K} v_{ijk} x_{ijk}
+
\mu
\left(
B-
\sum_{i=1}^{I} \sum_{j=1}^{J_i} \sum_{k=1}^{K} q_k x_{ijk}
\right).
\label{eq:max_X}
\end{equation}

The corresponding dual function is
\begin{equation}
\phi(\mu)=\max_{X}\mathcal{L}(X,\mu),
\end{equation}
and the dual problem is $\min_{\mu\ge 0}\phi(\mu)$.

For a fixed $\mu$, the optimization over $X$ decomposes across account-photo pairs. Specifically, for each pair $(i,j)$, we solve
\[
\max_{\{x_{ijk}\}_{k=1}^{K}}
\sum_{k=1}^{K}(v_{ijk}-\mu q_k)x_{ijk}
\quad
\text{s.t. }
\sum_{k=1}^{K}x_{ijk}=1,\;
0\le x_{ijk}\le 1.
\]
Therefore, the optimal quota choice for pair $(i,j)$ is
\begin{equation}
k^*(i,j;\mu)=\arg\max_{k}\big(v_{ijk}-\mu q_k\big),
\label{eq:kstar}
\end{equation}
and the corresponding solution is one-hot:
\[
x_{ijk^*}=1,\qquad x_{ijk}=0 \ \text{for } k\neq k^*.
\]
Hence, although the problem is relaxed, the optimal solution to each Lagrangian subproblem remains integral.

Define the capacity residual function
\begin{equation}
g(\mu)=\sum_{i=1}^{I}\sum_{j=1}^{J_i}\sum_{k=1}^{K} q_k x_{ijk}(\mu)-B,
\label{eq:glambda}
\end{equation}
where $x_{ijk}(\mu)$ is the optimal solution induced by Eq.~\ref{eq:kstar}. The following proposition shows that $g(\mu)$ is monotone.

\begin{proposition}
$g(\mu)$ is monotone non-increasing with respect to $\mu$.
\end{proposition}

\begin{proof}
Fix any account-photo pair $(i,j)$ and omit the indices $(i,j)$ for brevity. Let $k_1$ and $k_2$ be the optimal quota indices under dual variables $0<\mu_1<\mu_2$, respectively. By optimality,
\[
v_{k_1}-\mu_1 q_{k_1}\ge v_{k_2}-\mu_1 q_{k_2},
\]
and
\[
v_{k_2}-\mu_2 q_{k_2}\ge v_{k_1}-\mu_2 q_{k_1}.
\]
Adding the two inequalities yields
\[
(\mu_2-\mu_1)(q_{k_1}-q_{k_2})\ge 0.
\]
Since $\mu_2>\mu_1$, we obtain $q_{k_1}\ge q_{k_2}$. Therefore, the selected quota for each pair is non‑increasing in $\mu$. Summing over all pairs gives \(g(\mu_1)\ge g(\mu_2)\).

which proves that $g(\mu)$ is monotone non-increasing.
\end{proof}

By complementary slackness, the optimal dual variable $\mu^*$ satisfies
\begin{equation}
\mu^*\, g(\mu^*)=0.
\label{eq:kkt}
\end{equation}
Because $g(\mu)$ is monotone non-increasing, we can efficiently search for $\mu^*$ via bisection. Following this idea, we adopt a dual-based binary search solver (DBSSolver)~\cite{zhao2019unified} to obtain a near-optimal large-scale allocation solution.

\section{Experiments}

\subsection{Dataset}

We evaluate ACQ on a real-world dataset collected from Kuaishou's large-scale intelligent advertising delivery platform. Each sample corresponds to an account-photo pair under a given creative quota, with the target defined as the observed ad revenue over the evaluation window. The dataset contains approximately 60 million samples. As shown in Table~\ref{tab:dataset}, the target distribution is highly skewed, with a median of 0 and a long positive tail, which is consistent with the zero-heavy ad-revenue prediction setting studied in this paper.

\begin{table}[h!]
\caption{Dataset statistics.}
\label{tab:dataset}
\centering
\small
\setlength{\tabcolsep}{5pt}
\renewcommand{\arraystretch}{1.0}
\begin{tabular}{|c|c|c|c|c|c|}
\hline
Data Size & Mean & Median & Std. & Min & Max \\
\hline
60M & 0.81 & 0.00 & 40.10 & 0.00 & 20000.00 \\
\hline
\multicolumn{2}{|c|}{Sparse features} & \multicolumn{4}{p{8.2cm}|}{product name, first industry name, second industry name} \\
\hline
\multicolumn{2}{|c|}{Dense features} & \multicolumn{4}{p{8.2cm}|}{historical ad revenue, target ad revenue, conversion, and impression statistics over multiple time windows} \\
\hline
\end{tabular}
\end{table}

\subsection{Offline Experiments}

We conduct offline evaluation of ACQ for both the prediction and allocation modules using production logs, where training and validation are performed on temporally consecutive data slices.

\paragraph{Baselines.}
\begin{itemize}
    \item \textbf{UBTM}: Our proposed model based on an unbalanced binary tree.
    \item \textbf{DNN}: A standard deep regression model trained with mean squared error.
    \item \textbf{ZILN}~\cite{wang2019deep}: A deep model trained with the zero-inflated log-normal loss, which is widely used for zero-heavy value prediction.
    \item \textbf{CREAD}~\cite{sun2024cread}: A discretization-based method that transforms continuous labels into multiple intervals, trains multiple binary classifiers, and reconstructs the final prediction.
\end{itemize}

\paragraph{Evaluation metrics.}
\begin{itemize}
    \item \textbf{MSE}: Measures the average squared error between predicted and observed ad revenue.
    \item \textbf{AUC}: Evaluates the model's ability to separate zero-ad-revenue samples from positive-ad-revenue samples.
    \item \textbf{PAUC}: Measures ranking quality within the positive-ad-revenue subset, which is more sensitive to the ordering of valuable samples.
    \item \textbf{GAUC}: A grouped AUC weighted by account-level ad revenue, which emphasizes performance on high-value accounts.
\end{itemize}

\begin{table}[t]
  \centering
  \caption{Offline prediction results on the real-world dataset.}
  \label{tab:competitive_analysis}
  \begin{tabular}{llcccc}
    \toprule
    Main Method & Variant & AUC$\uparrow$ & MSE$\downarrow$ & PAUC$\uparrow$ & GAUC$\uparrow$ \\
    \midrule
    ZILN \cite{wang2019deep}   & -       & 0.927180 & NaN*   & 0.240395 & 0.653189 \\
    CREAD \cite{sun2024cread} & -       & 0.806779 & 980.21 & 0.573294 & 0.623219 \\
    \midrule
    \multirow{4}{*}{DNN}      & control & 0.884369 & 820.91 & 0.740429 & 0.689695 \\
                              & mon     & 0.904987 & 777.41 & 0.741367 & 0.678157 \\
                              & sub     & 0.905742 & 885.67 & 0.741194 & 0.682286 \\
                              & smo     & 0.906678 & 821.62 & 0.740722 & 0.682286 \\
    \midrule
    \multirow{4}{*}{UBTM (ours)} & control & 0.926627 & 812.98 & 0.745376 & 0.682656 \\
                                 & mon     & 0.930778 & 852.23 & 0.736744 & 0.675737 \\
                                 & sub     & $\boldsymbol{0.933521}$ & 783.63 & $\boldsymbol{0.745545}$ & $\boldsymbol{0.689926}$ \\
                                 & smo     & 0.924065 & $\boldsymbol{749.76}$ & 0.725943 & 0.663382 \\
    \bottomrule
  \end{tabular}
  \begin{minipage}{1.05\textwidth}
    \centering
    \footnotesize * Extremely large predictions lead to numerical overflow when computing squared errors.
  \end{minipage}
\end{table}

\begin{figure}[t!]
    \centering
    \includegraphics[width=0.98\linewidth]{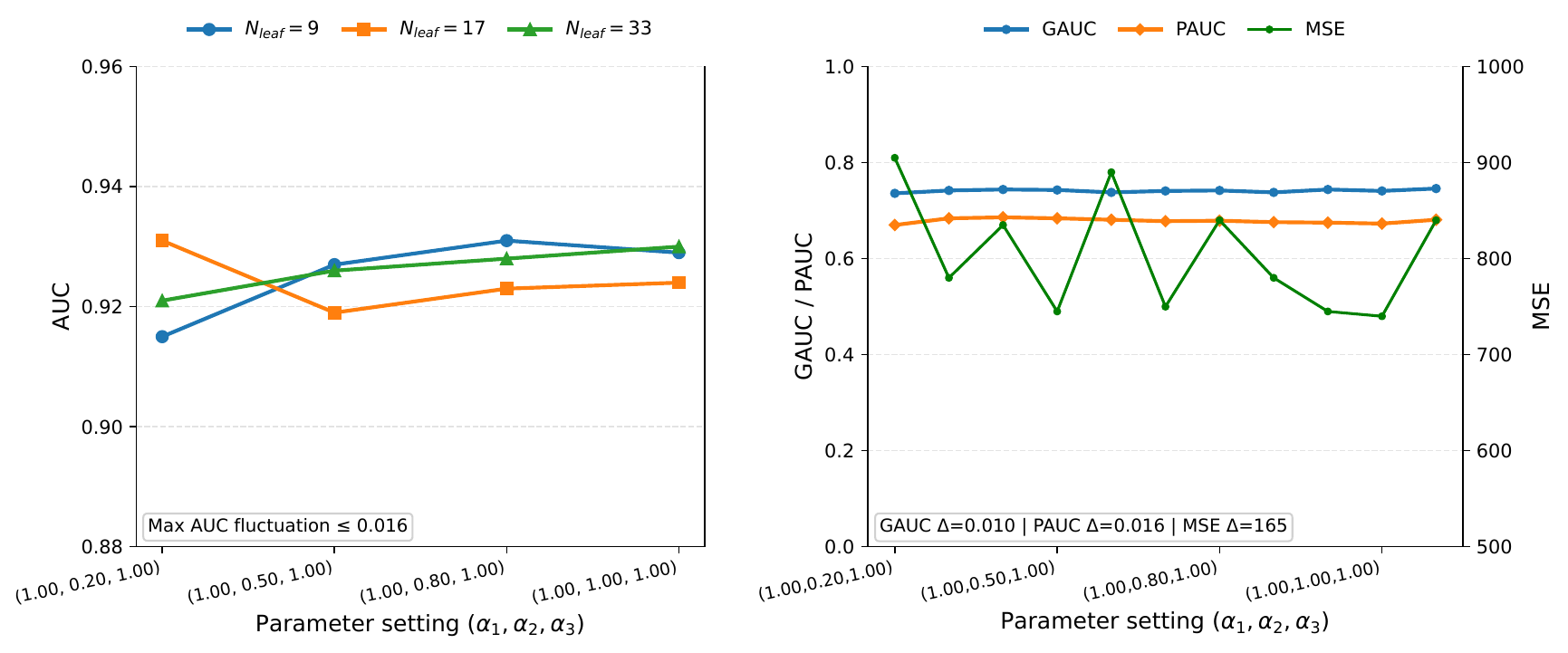}
    \caption{Sensitivity analysis of ACQ under different loss weights and tree sizes.}
    \label{fig:parameter_sensitivity}
\end{figure}

\begin{table*}[t]
  \centering
  \caption{Solver efficiency comparison between GLPK and the DBSSolver.}
  \label{tab:solver_unified}
  \setlength{\tabcolsep}{10pt}
  \begin{tabular}{lcccc}
    \toprule
    Dataset Size & \multicolumn{2}{c}{GLPK} & \multicolumn{2}{c}{DBSSolver(ours)} \\
    \cmidrule(lr){2-3} \cmidrule(lr){4-5}
    & Dual ($\times10^{-4}$) & Time (min) & Dual ($\times10^{-4}$) & Time (min) \\
    \midrule
    100k & 6.3316 & 69.6    & 6.3324 & 1.83 \\
    200k & 6.3312 & 312.0   & 6.3324 & 1.84 \\
    500k & 6.3305 & 2174.4  & 6.3324 & 1.94 \\
    \bottomrule
  \end{tabular}
\end{table*}

\begin{table*}[t]
  \centering
  \caption{Offline allocation results.}
  \label{tab:Offline_allocation_results}
  \makebox[\textwidth][c]{
    \begin{tabular}{lcccccc}
      \toprule
      Date & Ad revenue uplift & Creatives uplift & ACQ ad revenue & Rule-based ad revenue & ACQ creatives & Rule-based creatives \\
      \midrule
      Day1 & \textbf{22.72\%} & 0.04\% & \textbf{27,927} & 22,756 & \textbf{59,061} & 59,039 \\
      Day2 & \textbf{16.80\%} & 0.20\% & \textbf{33,449} & 28,638 & \textbf{58,355} & 58,242 \\
      Day3 & \textbf{18.64\%} & -0.25\% & \textbf{20,355} & 17,157 & \textbf{58,923} & 59,073 \\
      Day4 & \textbf{27.23\%} & 0.04\% & \textbf{25,687} & 20,190 & \textbf{62,671} & 62,645 \\
      Day5 & \textbf{26.51\%} & -0.13\% & \textbf{30,893} & 24,420 & \textbf{64,422} & 64,506 \\
      Day6 & \textbf{10.61\%} & 0.30\% & \textbf{18,879} & 17,064 & \textbf{64,891} & 64,698 \\
      Day7 & \textbf{11.96\%} & 0.02\% & \textbf{21,957} & 19,611 & \textbf{69,332} & 69,319 \\
      Day8 & \textbf{16.30\%} & -0.40\% & \textbf{29,123} & 25,041 & \textbf{74,511} & 74,809 \\
      \bottomrule
    \end{tabular}
  }
  \begin{center}
    \begin{minipage}{0.85\textwidth}
      \footnotesize \textit{Note:} All ad revenue and creative counts are reported in thousands (k).
    \end{minipage}
  \end{center}
\end{table*}

\paragraph{Experimental conclusions.}

Table~\ref{tab:competitive_analysis} reports the offline performance of the prediction module. Here, ``mon'', ``sub'', and ``smo'' denote the incorporation of monotonicity, submodularity, and smoothness, respectively, while ``control'' denotes the base model without these structural designs.

Several observations can be drawn. First, UBTM consistently outperforms the non-tree baselines on most ranking-oriented metrics, which validates the advantage of hierarchical target modeling for highly skewed ad-revenue prediction. In particular, \textbf{UBTM-sub} achieves the best AUC, PAUC, and GAUC, suggesting that modeling monotonicity together with diminishing marginal returns is beneficial for identifying high-value samples. Second, compared with the DNN variants, UBTM generally provides a better balance between regression accuracy and ranking quality, indicating that the unbalanced binary tree offers a more suitable inductive bias for zero-heavy and long-tailed ad-revenue data. Third, the smoothness constraint improves MSE for UBTM but degrades ranking-oriented metrics, suggesting that smoothness regularization may improve numerical fit while being less aligned with the downstream allocation objective.

Table~\ref{tab:solver_unified} compares the proposed DBSSolver with GLPK. DBSSolver achieves nearly identical dual values while reducing runtime by more than an order of magnitude, and its advantage becomes more pronounced as the problem size increases. These results confirm that the proposed solver is suitable for large-scale deployment.

Table~\ref{tab:Offline_allocation_results} presents the offline allocation results. Under an almost identical total number of allocated creatives, ACQ consistently yields higher ad revenue than the rule-based strategy across all evaluation days. This shows that the gain mainly comes from better quota allocation rather than simply assigning more creatives. The ad-revenue uplift ranges from 10.61\% to 27.23\%, with an average improvement of 18.08\%.

\paragraph{Parameter sensitivity.}

Figure~\ref{fig:parameter_sensitivity} shows that UBTM is relatively robust to the number of leaf nodes in the unbalanced binary tree and to the choice of multi-task loss weights. The performance varies only moderately across a reasonable range of hyperparameter settings.

\subsection{Online Experiments}

\subsubsection{Online Framework}

\begin{figure*}[t]   
    \centering   
    \includegraphics[width=1\linewidth]{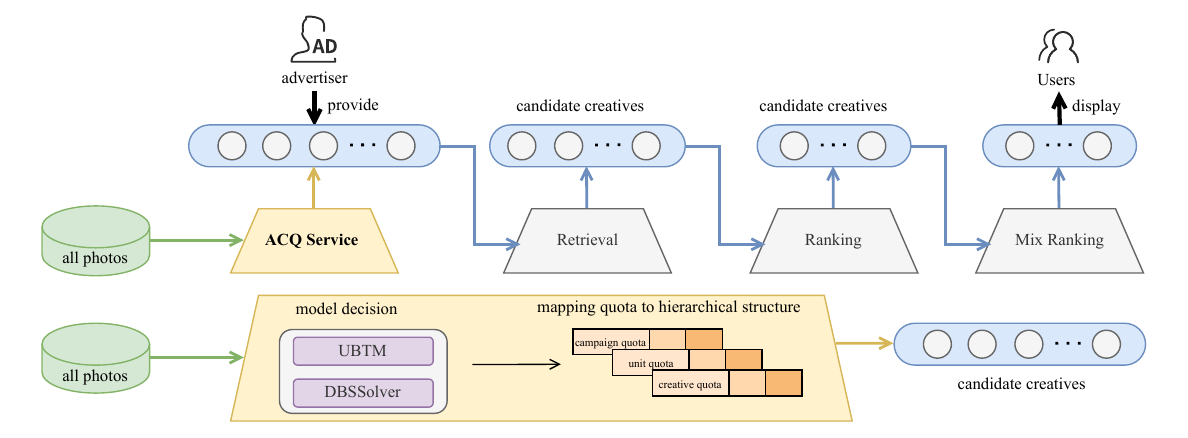}   
    \caption{Offline-to-online serving workflow of ACQ.}  \label{fig:structure_online.drawio.png} 
\end{figure*}

Figure~\ref{fig:structure_online.drawio.png} shows the online serving workflow of ACQ. Offline, UBTM and the DBSSolver determine creative quotas for account-photo pairs, which are then mapped to the hierarchical campaign--unit--creative structure and stored in Redis. Online, the advertising system retrieves these allocation results to support retrieval, ranking, and mix-ranking for candidate creative generation and delivery.

\subsubsection{A/B Test Results}

The ACQ framework was deployed on Kuaishou's advertising platform, where a 30-day A/B test was conducted to evaluate its effectiveness.. The A/B test results show a 6.20\% increase in actual ad revenue, a 4.46\% reduction in the total number of creatives, a 14.86\% increase in active creatives, a 0.15\% decrease in the total number of photos, and a 6.24\% increase in active photos. These results indicate that the ACQ framework optimizes resource allocation, improving cost efficiency by reducing redundant creatives while enhancing the proportion of active creatives and photos.

\section{Conclusion}

In this paper, we proposed ACQ, a two-stage framework for the creative quota allocation problem in digital advertising. ACQ aims to determine how many creatives should be generated from each photo so as to improve total ad revenue under platform capacity constraints. In the first stage, we developed UBTM, a multi-task prediction model based on an unbalanced binary tree, to address the highly skewed ad-revenue distribution while incorporating desirable structural properties such as monotonicity and submodularity. In the second stage, we formulated quota allocation as a large-scale optimization problem and designed an efficient dual-based solver, DBSSolver, to obtain practical solutions for industrial deployment. Extensive offline experiments and online A/B tests on Kuaishou's advertising delivery platform demonstrated the effectiveness of ACQ, showing that it improves ad revenue while enabling more efficient creative allocation at scale.

%
%
%
\bibliographystyle{splncs04}
\bibliography{sample-base}
\end{document}